\documentclass{article}

\PassOptionsToPackage{numbers, compress}{natbib}

\usepackage[final]{nips_2018}

\usepackage[utf8]{inputenc} 
\usepackage[T1]{fontenc}    
\usepackage{hyperref}
\usepackage{cleveref}
\usepackage{url}            
\usepackage{booktabs}       
\usepackage{amsfonts}       
\usepackage{nicefrac}       
\usepackage{microtype}      

\usepackage{dsfont}
\usepackage[numbers]{natbib}
\usepackage{xcolor}
\usepackage{amsmath}
\usepackage{amssymb}
\usepackage{comment}
\usepackage{floatrow}
\usepackage{subfigure}
\usepackage{blindtext}
\usepackage{color,soul}
\usepackage[final]{graphicx}

\newcommand{\Lagr}{\mathcal{L}}

\newcommand{\ie}{\textit{i}.\textit{e}., }
\newcommand{\eg}{\textit{e}.\textit{g}. }
\newcommand{\FM}{$\text{F}_{1}\text{-score }$}
\newcommand{\err}{\epsilon}
\newfloatcommand{capbtabbox}{table}[][\FBwidth]
\DeclareMathOperator*{\argmin}{arg\,min\;}

\title{A Hybrid Instance-based Transfer Learning Method}

\author{
  Azin Asgarian\\
  Georgian Partners Inc\\
  \And
    Parinaz Sobhani\\
    Georgian Partners Inc\\
  \And
    Ji Chao Zhang\\
    Georgian Partners Inc\\
  \And
    Madalin Mihailescu\\
    Georgian Partners Inc\\
  \And
    Ariel Sibilia\\
    Cority Inc\\
  \And
    Ahmed Bilal Ashraf\\
    University of Manitoba\\
  \And
    Babak Taati\\
    University Health Network\\
}

\begin{document}
\maketitle
\vspace{-.2cm}
\begin{abstract}
\vspace{-.2cm}
In recent years, supervised machine learning models have demonstrated tremendous success in a variety of application domains. Despite the promising results, these successful models are data hungry and their performance relies heavily on the size of training data. However, in many healthcare applications it is difficult to collect sufficiently large training datasets. Transfer learning can help overcome this issue by transferring the knowledge from readily available datasets (source) to a new dataset (target). In this work, we propose a hybrid instance-based transfer learning method that outperforms a set of baselines including state-of-the-art instance-based transfer learning approaches. Our method uses a probabilistic weighting strategy to fuse information from the source domain to the model learned in the target domain. Our method is generic, applicable to multiple source domains, and robust with respect to negative transfer. We demonstrate the effectiveness of our approach through extensive experiments for two different applications.
\end{abstract}

\vspace{-.3cm}
\section{Introduction}
\vspace{-.3cm}
Transfer learning techniques attempt to improve the generalization capabilities of predictive models to a new domain (target domain) by leveraging knowledge learned from pre-existing domains (source domain). The underlying assumptions are that first, the target training data is not sufficiently expressive, and second, that the source and target domains have some similarities, but they are not identical. In this study, we focus on applying transfer learning to two different applications of great importance in healthcare: facial expression recognition and injury prediction.

In recent years, facial expression recognition has been widely used to solve many challenges in healthcare, including increasing drivers' safety~\cite{McDuff_2013_CVPR_Workshops}, improving the quality of mental health~\cite{mcclure2003facial},  pain monitoring and management ~\cite{ashraf2009painful}, and medication adherence~\cite{hanina2012method}. Injury prediction is also of great importance in trying to prevent occupational accidents, which is a major problem in many industries~\cite{ILO, sarkaretal, rossi2017effective}. As reported by the International Labour Organization (ILO), there are more than 374 million work-related injuries and illnesses reported at work every year~\cite{ILO}. Effective transfer learning techniques could benefit both problems: when developing facial expression analysis models for healthcare applications, it is difficult to collect patient data (target domain), while data from healthy people is readily available. Similarly, accidents are rare events; so it is difficult to collect enough training data from one company/industry and it is desirable to effectively fuse information from multiple sources. 

Despite the advances in these fields, the generalization capabilities of models trained for these tasks to new domains is still challenging~\cite{Bayesian_AAM, asgarian2017subspace, haase2014instance}. To address this issue, we propose an instance-based transfer learning method which is a generalization of the method presented in~\cite{huang2007correcting}. We give an overview of our approach in Section~\ref{sec:method}. Unlike previous weighting methods~\cite{haase2014instance} that only measure the similarity to the target domain, our hybrid weighting strategy also considers the effectiveness of source samples in the target task. Our method assigns soft weights to source samples and, therefore, avoids hard decisions as done in~\cite{gong2013connecting}. We evaluate the effectiveness of our model on the two described tasks in Section \ref{sec:experiments}.

\section{Method}
\vspace{-.38cm}
\label{sec:method}
Given a loss function $\Lagr(.)$ and a small set of target training samples \mbox{ $\left\{(x_i,y_i)| i \in \left\{1,2, ...,N_T\right\}\right\}$,} the goal of supervised learning is to find model $\mathcal{A}^*$ that minimizes the expected error, \ie \mbox{$\mathcal{A}^* = \underset{\mathcal{A}\in\mathbb{A}}\argmin \mathbb{ E}_{x\sim P_T}\big[\Lagr(\mathcal{A}(x), y)\big]$.} Here $P_T$ is the probability distribution of target samples and $x$ is an arbitrary sample. To transfer the knowledge from source domain ($S$) to target domain ($T$), we follow the idea of importance sampling~\cite{liu2008monte}. Assuming an infinite number of training samples, we can express the expected error based on target and source samples as follows~\cite{haase2014instance}. 
\small
\begin{equation}
\begin{split}
\label{eq:2}
\mathbb{E}_{x\sim P_T}\big[\underbrace{\Lagr(\mathcal{A}(x), y)}_{\err(x)}\big] &= \int\err(x) P_T(x)dx = \int\err(x) \underbrace{\left[ \alpha + (1-\alpha)\frac{P_S(x)}{P_S(x)} \right]}_{=1} P_T(x)dx \\
          &= \alpha\mathbb{E}_{x\sim P_T}\Big[\err(x)\Big] + (1-\alpha)\mathbb{E}_{x\sim P_S}\Biggl[\err(x)\underbrace{\frac{P_T(x)}{P_S(x)}}_{w_x}\Biggr]
\end{split}
\vspace{-.07cm}
\end{equation}
\normalsize
Here $\err(x)$ is the error for each sample and $\alpha$ is a hyper-parameter that controls the overall relative importance between source and target samples. Source sample weights \mbox{$\{w_{x_j}=\frac{P_T(x_j)}{P_S(x_j)}\mid j\in\{1, ...,N_S\}\}$} have a key role in instance-based transfer learning methods, as they control the individual effect of source samples. Considering the case where a finite number of source ($N_S$) and target ($N_T$) training samples are available, we can replace the expected values and other terms in Equation~\ref{eq:2} with their respective counterparts.
\small
\begin{equation}
\begin{aligned}
\label{eq:3}
\Theta^* = \underset{\Theta}{\arg\min}\Biggl(\frac{\alpha}{N_T}\sum_{i=1}^{N_T} \epsilon(x_i,\Theta) + \frac{1-\alpha}{N_S}\sum_{j=1}^{N_S} \epsilon(x_j,\Theta)w_{x_j} \Biggr)
\end{aligned}
\end{equation}
\normalsize
The source-only and target-only models ($\mathcal{A}_S$ and $\mathcal{A}_T$) serve as minimum baselines that a transfer learning approach must outperform. A useful transfer learning method should also outperform a model trained directly on the union of the source and target data denoted by \mbox{$\mathcal{A}_{S\cup T}$.} In the following, we describe three additional baselines included in our experimental analysis, and our contribution.

\vspace{-.05cm}
\textbf{Weights All One:} 
As a trivial baseline, we also have an instance-weighted model where all the weights are set to $\mathbf{1}$, \ie $\mathbf{W}_S = \mathds{1}$. We denote this model with $\mathcal{A}_\mathbf{1}$. This is similar to $\mathcal{A}_{S\cup T}$, except here we use $\alpha$ to determine the relative overall importance between source and target samples.

\vspace{-.05cm}
\textbf{Gaussian Weights:}
Another solution is to use a generative approach; we can assume normal distributions for target and source samples, which leads to \mbox{$w_{x_j} = \frac{P_T(x_j)}{P_S(x_j)} = \frac{\mathcal{N}(x_j;\mu_T,\Sigma_T)}{\mathcal{N}(x_j;\mu_S,\Sigma_S)}$}. Here $\mu_T$ and $\mu_S $ indicate the mean and $\Sigma_T$ and $\Sigma_S$ indicate the covariance matrices for target and source distributions respectively. We denote this model with $\mathcal{A}_G$.

\vspace{-.05cm}
\textbf{Jena Weights:} The current state-of-the-art instance-based method~\cite{haase2014instance} employs a heuristic weighting strategy. We only evaluate this method on facial expression recognition task as it is specifically developed for Active Appearance Models(AAMs). We call this model $\mathcal{A}_{Jena}$.

\vspace{-.05cm}
\textbf{Hybrid Weights (our contribution):}
Previous methods evaluate the weights associated to source samples only based on their similarity to the target domain. We argue that it is also important to measure the relevance of these samples to the target task. Therefore, we propose a weighting strategy that considers both of these factors. We define weights $w_{x}=w_{domain_x} + w_{task_x}$, where $w_{domain_x}$ measures the similarity of an arbitrary source sample $x$ to the target domain, while $w_{task_x}$ measures the importance of sample $x$ in the target task.

\vspace{-.03cm}
Given the assumption that only few target samples are available,  estimating  $P_T$ is an ill-posed problem. Therefore, for evaluating $w_{domain_x}$, instead of the generative approach of estimating $P_T$ and $P_S$, we directly approximate weights \mbox{$w_{domain_x} = \frac{P_T(x)}{P_S(x)}$} with a discriminative classifier.  Specifically, using \mbox{$\{(x, l_x)\ | x \in \text{source} \cup \text{target, and $l_x=1$ if $x \in$ source,  $l_x=0$ otherwise} \}$}, we train a binary classifer, e.g., logistic regression (LR), to differentiate source and target samples. We then use the learned weights of this classifier ($w_{lr}$ and $c_{lr}$) to calculate source sample weights \mbox{$w_{target_x}=\frac{P_T(x)}{P_S(x)}=\frac{1}{\exp(x^T w_{lr} + c_{lr})}$}. To obtain $w_{task_x}$, we train an instance of predictive model using all samples from source and target \mbox{($\mathcal{A}_{S\cup T}$).} We then use the uncertainty of this model about sample $x$ as $w_{task_x}$. In binary classification tasks, we define the uncertainty of a model to be the distance of sample $x$ to the decision boundary. Note that this value could be positive (thus adding to $w_{x}$) when the decision is correct, or negative (thus subtracting from $w_{x}$) when the decision is incorrect. In structured prediction tasks, we  use the reconstruction error. We call this model $\mathcal{A}_{HW}$. 

\vspace{-.3cm}
\section{Experiments}
\vspace{-.3cm}
\label{sec:experiments}
To evaluate our method, we conducted extensive experiments on two different tasks: facial expression recognition and injury prediction. We briefly describe them in the following. More examples and details are included in the supplementary materials (Section \ref{sec:supp}). 
\vspace{-.3cm}
\subsection{Facial Expression Recognition:}
\vspace{-.2cm}
We used 555 samples randomly selected from public datasets LFPW~\cite{belhumeur2013localizing}, Helen~\cite{le2012interactive},
CK+~\cite{ck_ex}, iBUG~\cite{sagonas2013300}, and AFW~\cite{zhu2012face}. Additionally, we selected 320 examples from the UNBC-McMaster Shoulder Pain Expression Archive~\cite{lucey2011painful} which contains real pain expressions from participants with shoulder injury. We considered two different settings: \mbox{\textbf{Setting 1: Real pain expressions}} where the UNBC-McMaster dataset was considered as the target domain and the rest of the datasets were considered as the source domain. \mbox{\textbf{Setting 2: Posed expressions}} where CK+ dataset was considered as the target domain. This setting is further challenging as target domain has multiple posed or fake expressions (\eg sadness, anger, etc.) that are absent in the source domain. In both settings, five examples were randomly selected from target for training, and the test set had 200 samples. To compare different models, we use two criteria as suggested in literature~\cite{zhu2012face,matthews2004active}. First, the normalized root mean square (RMS) error between the points of the predicted shape and the ground truth shape. Second, the percentage of test examples that converge to the ground truth shape given a tolerance in the RMS fitting error. We used Active Appearance Models (AAMs) as the base predictive model.

\begin{figure*}[htbp]
\vspace{-.3cm}
\centering
\subfigure[\footnotesize RMS fitting error (Setting 1)]{\label{fig:acc-a}\includegraphics[width=2.6in]{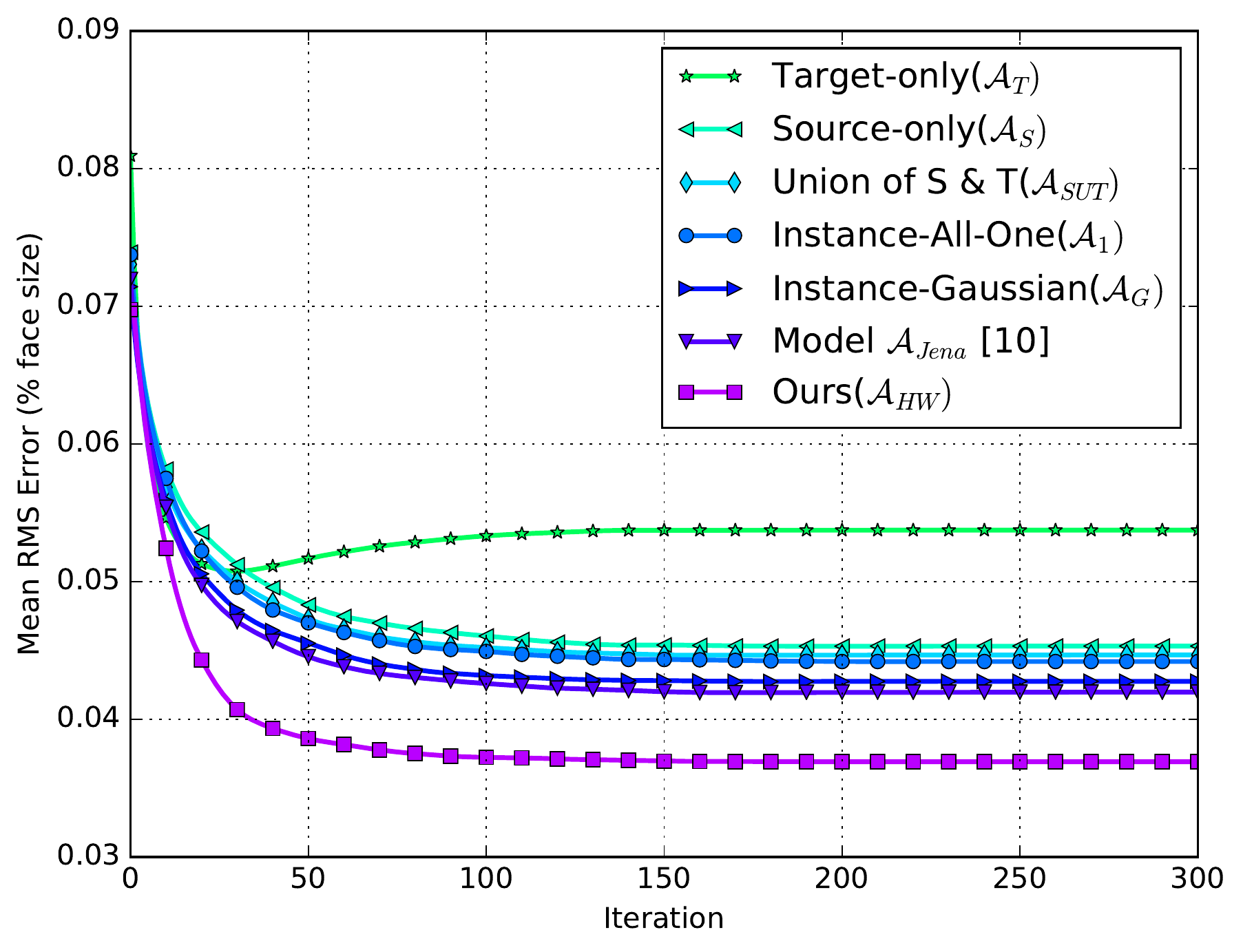}}
\vspace{-.2cm}
\hspace{.5cm}
\subfigure[\footnotesize \% of test examples converged (Setting 1)]{\label{fig:conv-a}\includegraphics[width=2.5in]{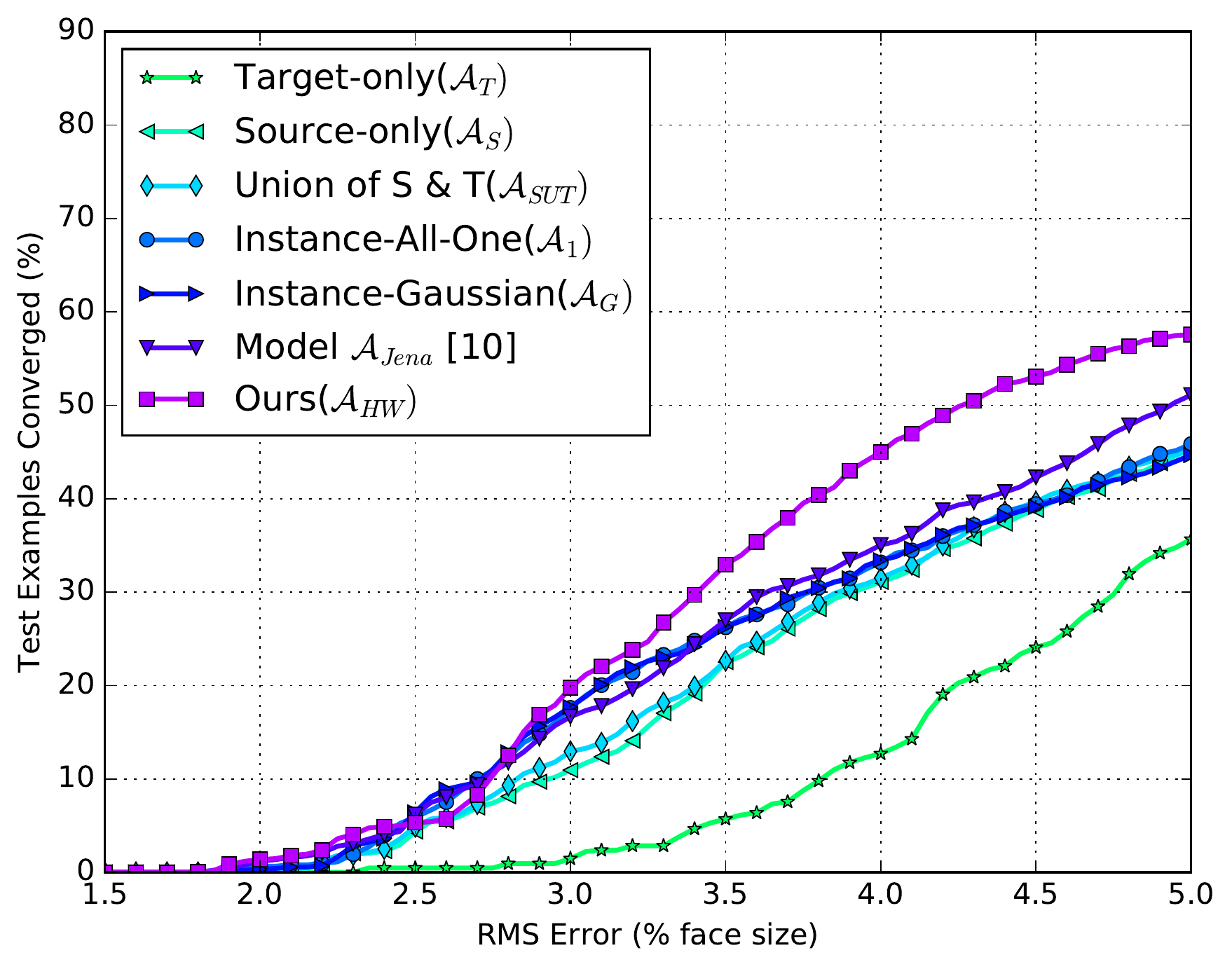}}
\vspace{-.25cm}
\subfigure[\footnotesize RMS fitting error (Setting 2)]{\label{fig:acc-b}\includegraphics[width=2.6in]{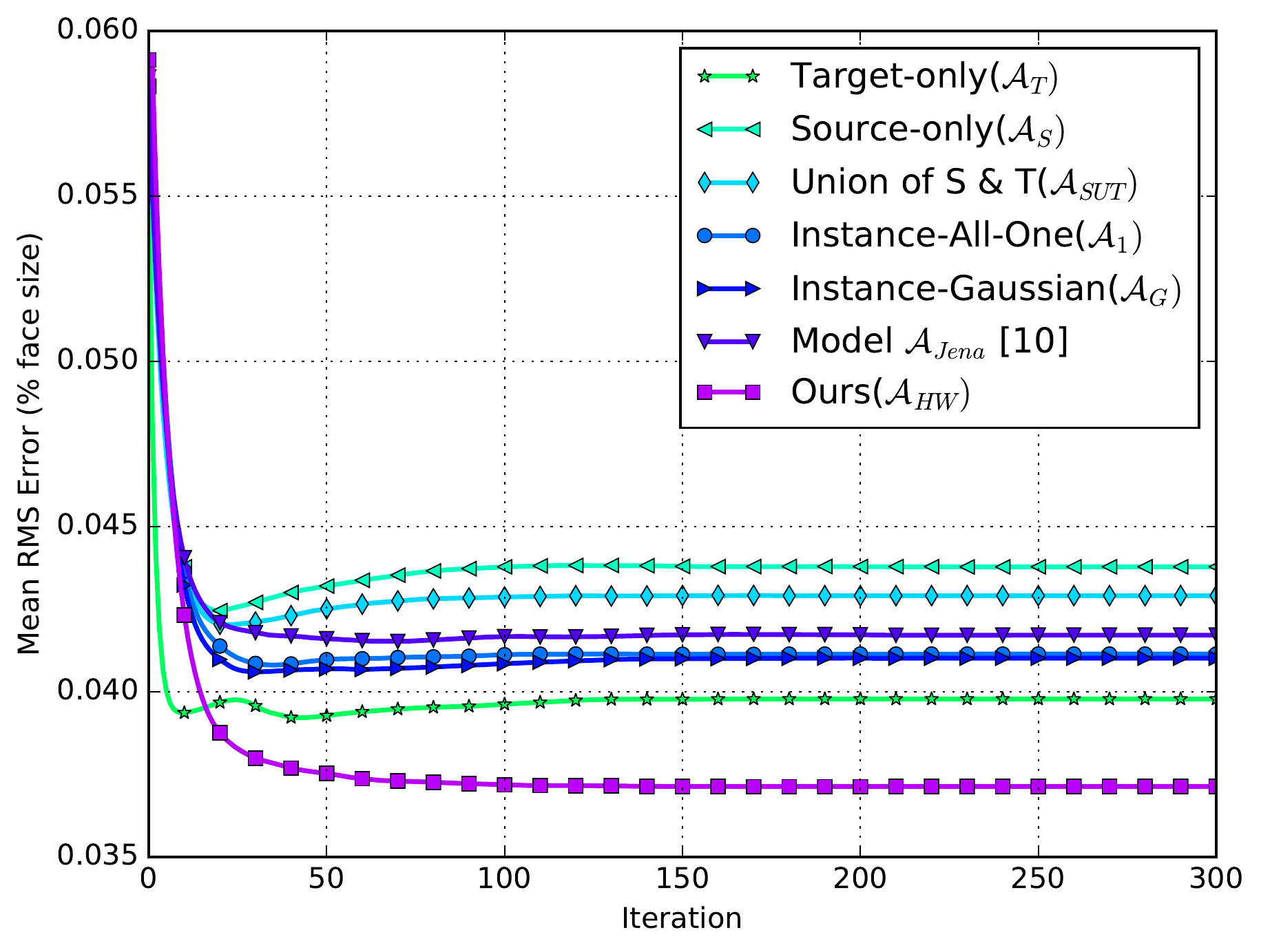}}
\hspace{.5cm}
\subfigure[\footnotesize \% of test examples converged (Setting 2)]{\label{fig:conv-b}\includegraphics[width=2.5in]{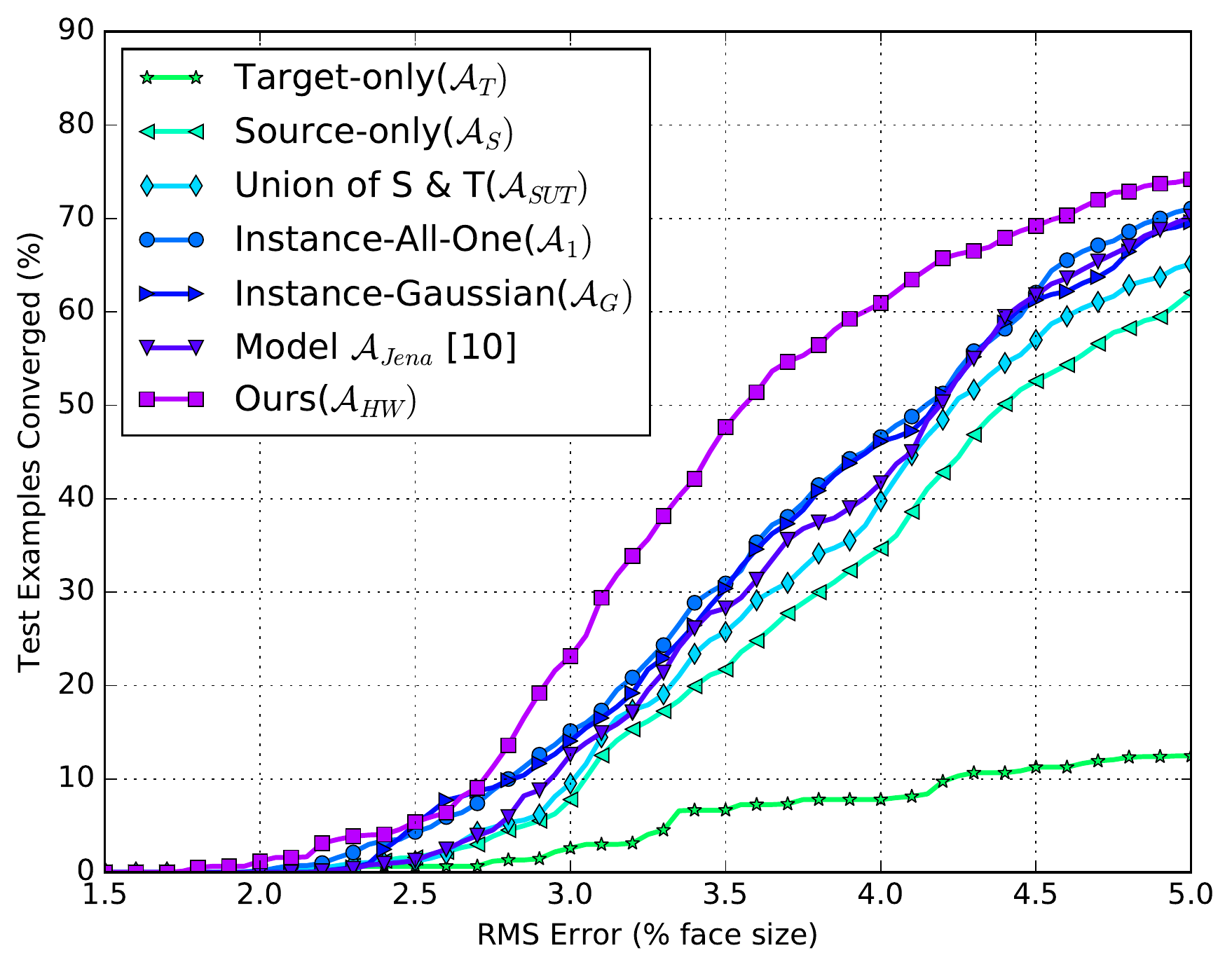}}
\caption{\small Comparison of RMS error and the percentage of converged test examples in setting 1 (UNBC-McMaster as target) and Setting 2 (CK+ as target).}
\label{Fig:curves}
\end{figure*}

\textbf{Results:}
We compare different models in terms of the RMS error and percentage convergence for  setting 1 (UNBC-McMaster as target) and Setting 2 (CK+ as target) in Figure \ref{Fig:curves}. The curves in Figures \ref{fig:acc-a} and \ref{fig:acc-b} show the RMS error averaged over converged test examples as a function of iterations. The plots in Figures \ref{fig:conv-a} and \ref{fig:conv-b} demonstrate the percentage of test examples that converged to the ground truth as a function of RMS error. For both settings, our approach outperforms all other methods in terms of RMS error as well as the percentage of test examples that converge to the ground truth. In our
approach, hyper-parameter $\alpha$ was set to 0.9 using cross-validation.

To compare different models, we have to consider fitting accuracy and coverage together. In setting 2, 
Figures \ref{fig:acc-b} and \ref{fig:conv-b}, the target-only model ($\mathcal{A}_{T}$) has a good fitting accuracy over converged trials, while the percentage of convergence is very low, most likely due to the lack of expressiveness of the model. On the other hand, the source-only model ($\mathcal{A}_{S}$) has a higher convergence rate, but the fitting accuracy is lower. Also model $\mathcal{A}_{S\cup T}$ performs slightly better than model $\mathcal{A}_{S}$ as a result of including target samples, highlighting the importance of transfer learning. The instance-weighted model~\cite{haase2014instance} ($\mathcal{A}_{Jena}$) has a small improvement in convergence over previous models, but unexpectedly performs worse than the target-only model in terms of the fitting accuracy. Our Hybrid model ($\mathcal{A}_{HW}$) improves the percentage of converged examples, and the fitting error is significantly decreased. 
\vspace{-.1cm}
\subsection{Injury Prediction:}
\vspace{-.2cm}
Two real-world datasets collected from Company-A's partners (Partner-1 and Partner-2) were used.\footnote{Company-A is a SaaS company that provides health and safety management services to organizations around the world. The names have been masked for blind review.} Data for both partners are represented by 38 engineered features that capture two groups of information per individual: general information (\eg age), and event-based information (\eg number of absences) collected during years 2016-2017. If an employee is injured in 2017, his/her record is labeled 1, otherwise as 0. In our transfer learning framework, we considered Partner-1's dataset as the target domain and Partner-2's dataset as the source domain. We trained all the models with 58,271 samples from target (12,225) and source (46,046) training sets, and evaluated them on 3,057 samples from target test set. Since the datasets were highly imbalanced (1-7\% injury cases), we used precision, recall, \FM (macro), and classification accuracy(CA) as our four evaluation metrics. In this experiment, we used XGBoost~\cite{friedman2001greedy} as the base predictive model.

\begin{figure}[htbp]
\vspace{-.2cm}
\begin{floatrow}
\capbtabbox{%
\begin{tabular}{ccccc}
\toprule
Method & Precision & Recall & $\text{F}_{1}\text{-score}$ & CA(\%)\\
\midrule
 $\mathcal{A}_{T}$              & 0.07 & 0.06 & 0.06 & \textbf{97}\\ 
 $\mathcal{A}_{S}$              & 0.04 & 0.18 & 0.07 & 91\\ 
 $\mathcal{A}_{S\cup T}$            & 0.13 & 0.06 & 0.08 & \textbf{97}\\ 
 $\mathcal{A}_\mathbf{1}$       & 0.06 & 0.12 & 0.08 & 95\\ 
 $\mathcal{A}_{G}$              & 0.07 & 0.16 & 0.10 & 95\\ 
 $\mathcal{A}_{HW}$            & 0.11 & 0.12 & \textbf{0.12} & \textbf{97}\\  
\bottomrule
\end{tabular}
\vspace{.15cm}
}{\caption{\small Performance of different methods on Partner-1's data.} \label{table:1}}
\ffigbox{%
  \includegraphics[width=0.7\linewidth]{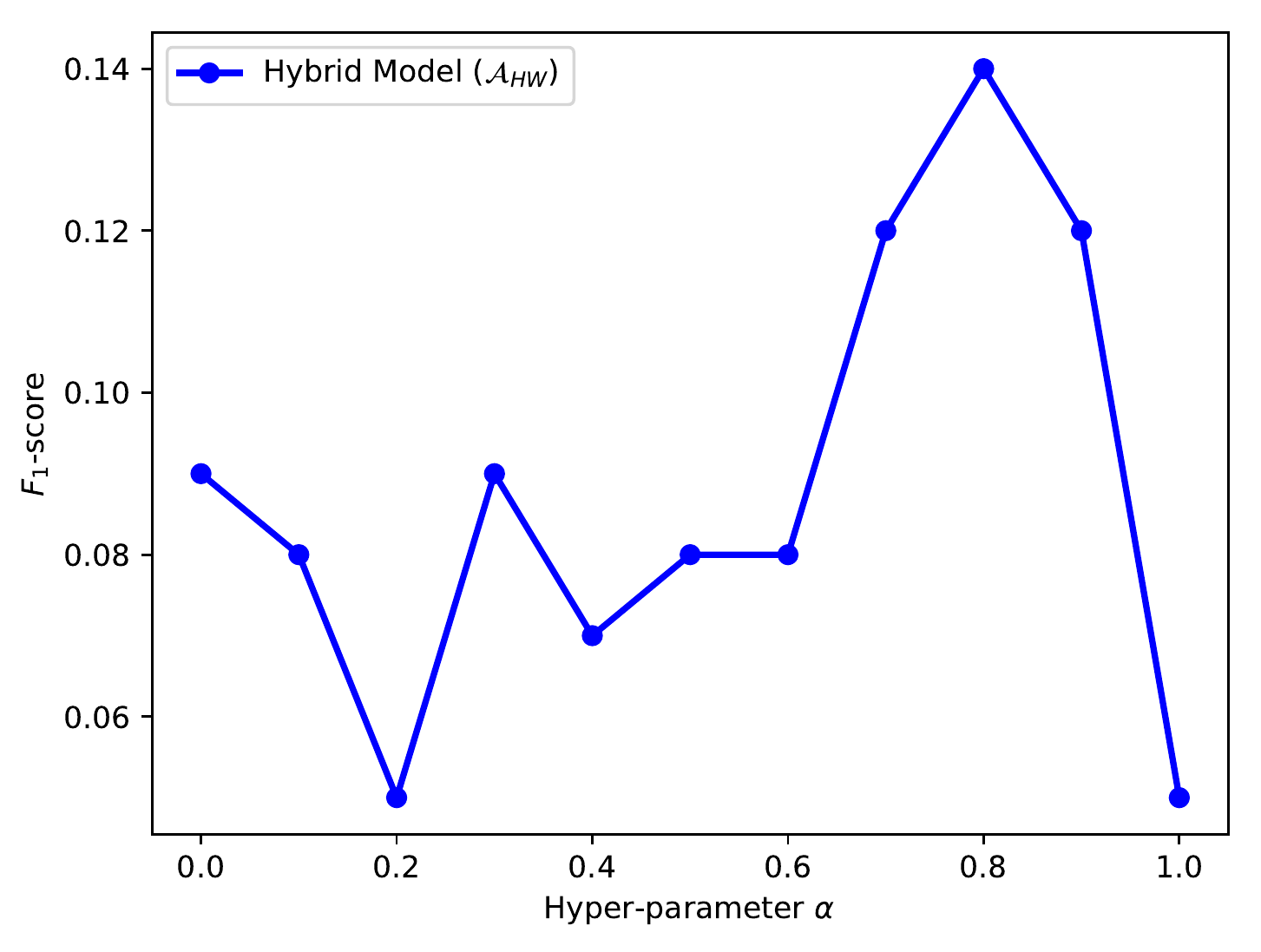}
  \vspace{-0.2cm}}{\caption{\small Effect of $\alpha$ on performance.} \label{fig:alpha}}
\end{floatrow}
\vspace{-.35cm}
\end{figure}

\textbf{Results:}
Results of our quantitative evaluation is shown in Table \ref{table:1}. We see that model $\mathcal{A}_T$ has a high classification accuracy, while the performance on injury class is very poor with \FM  equal to 0.06. This is possibly due to data sparsity issue and lack of expressiveness of the model. On the other hand, $\mathcal{A}_S$ has a higher \FM, but the precision and classification accuracy are diminished. Also model $\mathcal{A}_{S\cup T}$ performs better in terms of \FM and classification accuracy compared to both models $\mathcal{A}_T$ and $\mathcal{A}_S$. The best result is obtained with our model $\mathcal{A}_{HW}$, which increases the \FM significantly, while maintaining a high classification accuracy. In a separate analysis,\footnote{Included in the supplementary material} we noticed that model $\mathcal{A}_T$ only considers 13 features among the 38 provided features, most likely due to data sparsity issue. However, our model $\mathcal{A}_{HW}$ utilizes 23 features which includes $\mathcal{A}_{T}$'s 13 features plus 10 more features such as illness.
Figure \ref{fig:alpha} shows the $\text{F}_{1}\text{-score}$ obtained with model $\mathcal{A}_{HW}$ as a function of hyper-parameter $\alpha$. We can see that the best performance is achieved with $\alpha$ between 0.7 and 0.9. However, increasing or decreasing $\alpha$ results in lower \FM, as it enhances the influence of target or source samples. 

\section{Conclusion}
\vspace{-.2cm}
\label{sec:conclusion}
One of the biggest challenges of applying supervised machine learning models to healthcare applications is lack of sufficient training samples. In this paper, we proposed a hybrid instance-based transfer learning method to tackle this problem. We evaluate our method in two different settings, structured prediction with unstructured data and binary classification with structured data. We experimentally show that our approach improves the generalization capabilities of predictive models by leveraging knowledge from existing domain and outperforms all baselines including the state-of-the-art instance-based transfer learning approaches.
\small

\bibliographystyle{unsrtnat}
\bibliography{egbib}

\newpage
\section{Supplementary materials}
\label{sec:supp}
\begin{figure}[htbp]
\fboxsep=0mm
\fboxrule=2pt
\centering
\fcolorbox{blue}{white}{\includegraphics[width=\textwidth]{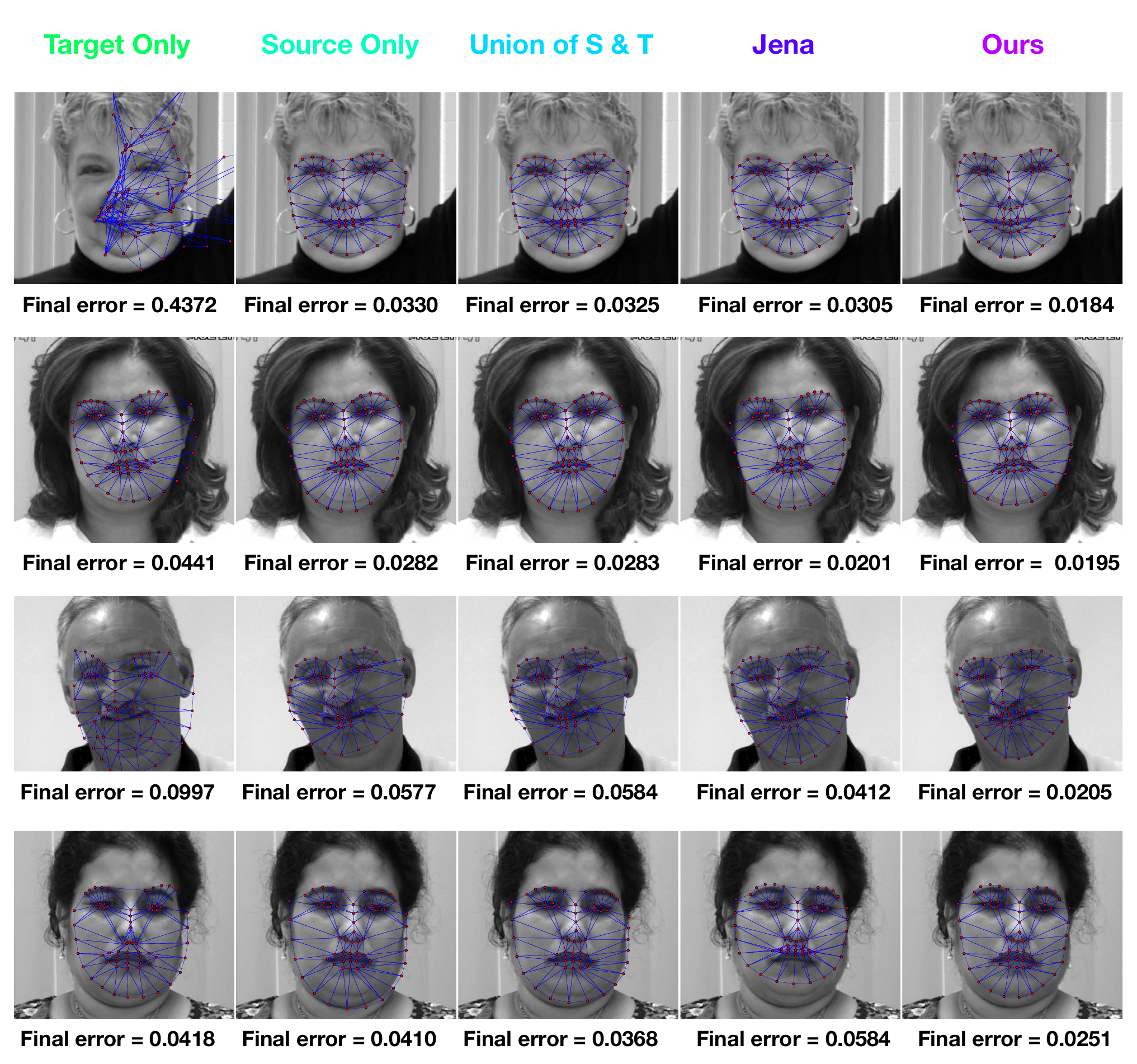}}
\caption{A visualization of AAM fitting and RMS error for 4 different test examples from UNBC-McMaster dataset obtained with different methods.}
\label{fig:aam}
\end{figure}

A visualization of AAM fitting results and RMS fitting error from setting 1 is shown in Figure \ref{fig:aam}. In this setting, the majority of images in the source domain are from young people
and children with happy or neutral expressions. However, in the target domain we have images from adults and seniors with pain expressions. As shown in the first column, the performance of model $\mathcal{A}_T$ trained only with few available examples from target domain is not desirable. Models $\mathcal{A}_S$ and $\mathcal{A}_{S\cup T}$ perform substantially better compared to target-only model with using the knowledge from source domain. Although these models can capture the features that are common between source and target (\eg structure of the face), they still lack the ability to capture specific features of target domain that are absent in source domain (\eg double chin, white eyebrows). An ideal transfer learning method should transfer the knowledge from source domain while preserving the target specific information. As shown in Figure \ref{fig:aam}, our model ($\mathcal{A}_{HW}$) performs considerably better than all the baselines as it can highlight the target-specific features while transferring the knowledge from source samples. Also, compared to model $\mathcal{A}_{Jena}$, our model is more robust. 

\begin{figure*}[htbp]
\centering
\subfigure[Features importance scores for model $\mathcal{A}_T$]{\label{fig:FI_T}\includegraphics[width=.49\textwidth]{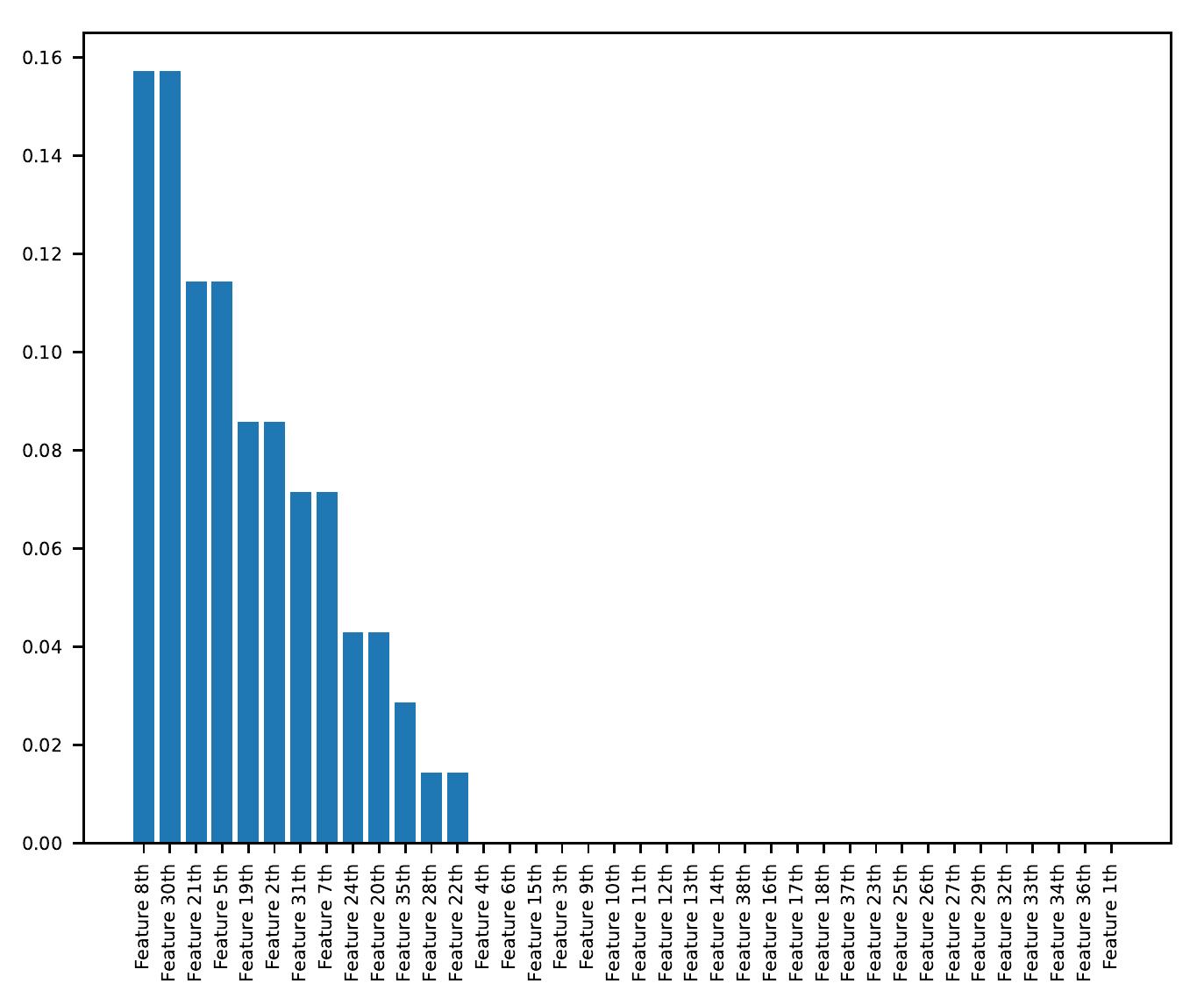}}
\subfigure[Features importance scores for model $\mathcal{A}_{HW}$]{\label{fig:FI_H}\includegraphics[width=.49\textwidth]{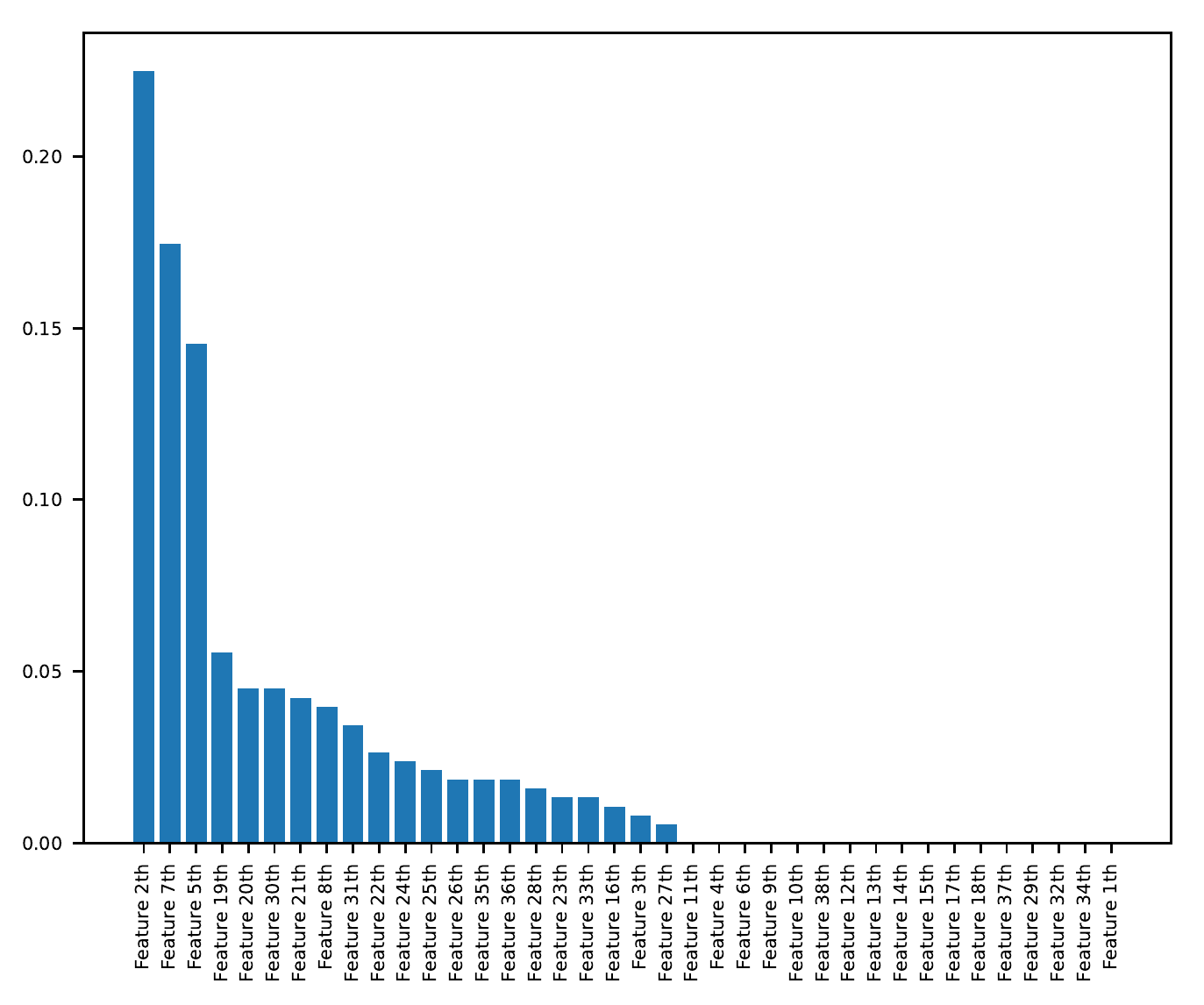}}
\hspace{.2cm}
\caption{Illustration of feature importance scores obtained with target-only model ($\mathcal{A}_T$) and our hybrid model ($\mathcal{A}_{HW}$). Feature names are masked for confidentiality reasons.}
\label{Fig:FI}
\end{figure*}

\newpage
Figure \ref{Fig:FI}: Feature importance score obtained with target-only model ($\mathcal{A}_T$) and our model ($\mathcal{A}_{HW}$) in injury prediction task. Among the 38 features provided in the training data, target-only model only utilizes 13 features. This is likely due to imbalance  and sparsity issues in data. However, our method uses 23 features. These 23 features include all the 13 features used in target-only model and 10 new features that seem to be intuitively important in injury prediction such as \texttt{number of illness}. 

\end{document}